# Measuring the Algorithmic Efficiency of Neural Networks

**Danny Hernandez** [*]

OpenAI

danny@openai.com

**Tom B. Brown**

OpenAI

tom@openai.com

## Abstract

Three factors drive the advance of AI: algorithmic innovation, data, and the amount of compute available for training. Algorithmic progress has traditionally been more difficult to quantify than compute and data. In this work, we argue that algorithmic progress has an aspect that is both straightforward to measure and interesting: reductions over time in the compute needed to reach past capabilities. We show that the number of floating-point operations required to train a classifier to AlexNet-level performance on ImageNet has decreased by a factor of 44x between 2012 and 2019. This corresponds to algorithmic efficiency doubling every 16 months over a period of 7 years. Notably, this outpaces the original Moore's law rate of improvement in hardware efficiency (11x over this period). We observe that hardware and algorithmic efficiency gains multiply and can be on a similar scale over meaningful horizons, which suggests that a good model of AI progress should integrate measures from both.

[*]Danny Hernandez led the research. Tom Brown paired on initial experiments, scoping, and debugging.

# Contents

# 1 Introduction

## 1.1 Measuring algorithmic progress in AI is critical to the field, policymakers, and industry leaders

There's widespread agreement there's been impressive progress in AI/ML in the domains of vision, natural language, and game playing in the last decade [Krizhevsky et al., 2012, Xie et al., 2016, Silver et al., 2018]. However, there's massive disagreement as to how much progress in capabilities we should expect in the near and long term [Grace et al., 2017]. For this reason, we believe measuring overall progress in AI/ML is a crucial question, because it can ground the discussion in evidence. Measuring AI progress is critical to policymakers, economists, industry leaders, potential researchers, and others trying to navigate this disagreement and decide how much money and attention to invest in AI.

For example, the compute used by the largest AI training runs per year grew at 300,000x between 2012 and 2018 [Amodei & Hernandez, 2018]. Given the divergence from the past trend of approximately Moore's Law level growth for such training runs, [Sastry et al., 2019] suggests policymakers increase funding for compute resources for academia, so they can continue to do the types of AI research that are becoming more expensive. Measurements of AI progress inform policymakers that are making such decisions.

Hardware trends are relatively quantifiable. Moore's Law explains much of the advance from mainframes, to personal computers, to omnipresent smartphones [Moore, 1965]. Better measurement of scientific progress has the potential for a lot of impact on a variety of fronts. Given the existing understanding of key hardware trends, we were primarily interested in measures that represented exclusively algorithmic improvement that could help paint a picture of the overall progress of the field.

We present measurements of algorithmic efficiency state of the arts over time that:

1. Are informative to a wide audience of decision makers
2. Help measure novel contributions produced with smaller amounts of compute

## 1.2 Efficiency is the primary way we measure algorithmic progress on classic computer science problems. We can apply the same lens to machine learning by holding performance constant

In a classic computer science problem like sorting, algorithmic quality is primarily measured in terms of how cost asymptotically increases with problem difficulty, generally denoted in Big O Notation. For example, quicksort [Hoare, 1962] has $O(n \log n)$ average cost in terms of operations to find a perfect solution whereas many sorting algorithms are $O(n^2)$ (where $n$ is the length of the list to be sorted). It's impractical to perform similar analysis for deep learning, because we're looking for approximate solutions and don't have as clear a measure of problem difficulty. For these reasons, in machine learning, algorithmic progress is often presented in terms of new states of the art, like a 1% absolute increase in top-5 accuracy on ImageNet, ignoring cost. It's difficult to reason about overall progress in terms of a large collection of such measures, because:

1. Performance is often measured in different units (accuracy, BLEU, points, ELO, cross-entropy loss, etc) and gains on many of the metrics are hard to interpret. For instance going from 94.99% accuracy to 99.99% accuracy is much more impressive than going from 89% to 94%.

2. The problems are unique and their difficulties aren't comparable quantitively, so assessment requires gaining an intuition for each problem.

3. Most research focuses on reporting overall performance improvements rather than efficiency improvements, so additional work is required to disentangle the gains due to algorithmic efficiency from the gains due to additional computation.

4. The benchmarks of interest are being solved more rapidly, which exacerbates 1) and 2). For instance it took 15 years to get to human-level performance on MNIST [LeCun et al., 1998], 7 years on ImageNet [Deng et al., 2009, Russakovsky et al., 2015], and GLUE [Wang et al., 2018] only lasted 9 months [Devlin et al., 2018, Liu et al., 2019].

We show that we can gain clear insights into efficiency trends by analyzing training costs while holding performance constant. We focused on training efficiency rather than inference efficiency, because we're more interested in what systems are possible to produce than how much it costs to run those systems. Though we note increased inference efficiency can have important economic implications [van den Oord et al., 2017]. In the research setting, we've typically found ourselves FLOPS bound rather than memory or communication bound. So we measured total floating-point operations used in training rather than parameters or another measure of efficiency.

We focused on AlexNet-level performance, which we measured as 79.1% top-5 accuracy on ImageNet. AlexNet kicked off the wave of interest in neural networks and ImageNet is still a benchmark of wide interest, so this measure provided a long running trend to analyze.

## 2 Related Work

### 2.1 Algorithmic progress had similar rate to Moore's Law in some domains over decades

Grace compared algorithmic progress to hardware progress looked at over several decades in the domains of chess, go, physics simulations, mixed integer programming, and SAT solvers [Grace, 2013]. Grace's overall conclusion was

> *Many of these areas appear to experience fast improvement, though the data are often noisy. For tasks in these areas, gains from algorithmic progress have been roughly fifty to one hundred percent as large as those from hardware progress. Improvements tend to be incremental, forming a relatively smooth curve on the scale of years*

For the most part, these estimates and their interpretation require substantial amounts of judgment. For instance, with chess and Go the approach was to use the available literature to estimate what kinds of returns came from a hardware doubling and then attribute all ELO improvement not explained by Moore's law to software. Additionally, Grace suggests we treat these estimates as "optimistic" rather than representative, because of increased saliency of problems that are making fast progress, problems with good measures being likely to progress faster, and the potential motivations of authors. Regardless, we think this related work shows that hardware and algorithmic progress can be on a similar scale, and that even a relatively simple model of progress should consider integrating measures from both domains.

Progress on mixed integer programming was particularly straightforward to measure, so we've extended the original analysis of that domain below [Bixby, 2012].

### 2.2 Linear programming gains were well-defined, steady, and faster than Moore's Law for 21 years

Unlike some other optimization domains Grace looked at, linear programming was of commercial interest for a long period. Progress is easy to track in this domain over this 21 year period because there were distinct releases of commercial software (CPLEX and Gurobi) that can be compared with hardware held fixed.

The trend of a 2x speedup every 13 months observed in Figure 1 is surprisingly consistent over a long time horizon. The smooth progress is partially explained by the measure being an aggregation of many problems of varying difficulty. Over this time Moore's Law yielded an efficiency gain of approximately 1500x.

**Caveats**

1. It's notable that the benchmark was designed and the analysis was performed by the CEO of Gurobi (a commercial MIPS solver) and that he had an incentive to demonstrate large amounts of progress.

2. It's worth pointing out the implications of the maximum search time of 30,000s for the optimal solution. When it took longer than 30,000s for the solver to find the optimal solution, 30,000s is what would be recorded. It's expected that the maximum search time would have been invoked more for earlier, weaker solvers. Thus, the maximum search time made earlier solvers look relatively stronger, making the overall estimate conservative for this benchmark. We think using a maximum search time is reasonable, but we expect the overall speedup is sensitive to it. In this sense, these measurements are a little different than the AlexNet accuracy measurements, where we waited for the capability to be demonstrated before measuring progress.

3. This is the related domain with highest amount of measured algorithmic efficiency progress we're aware of for this period of time.

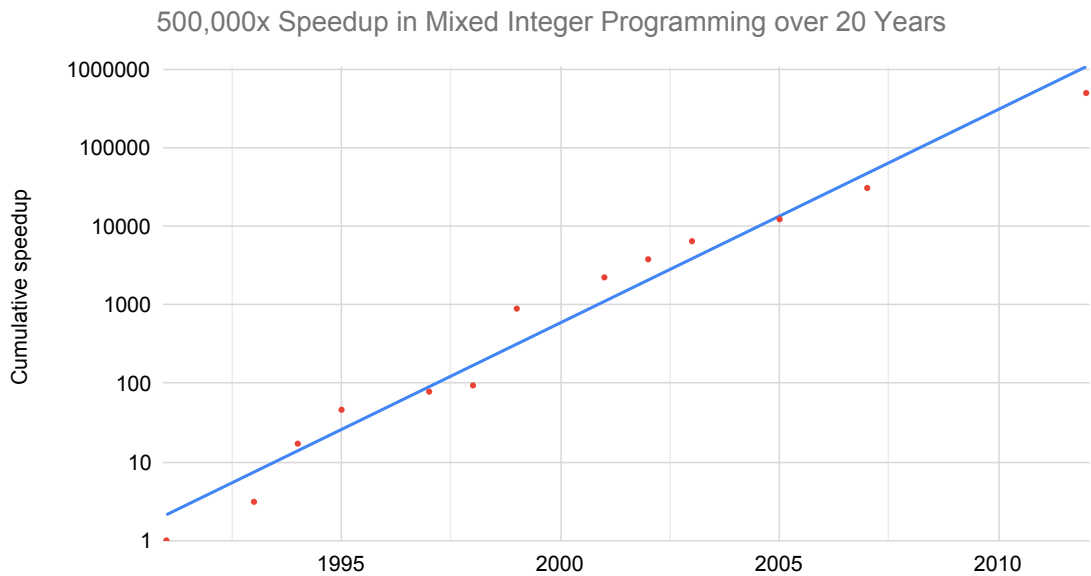

**Figure 1** A 2x speedup every 13 months was observed on a benchmark of 1,892 mixed-integer problems (MIPs), a subset of linear programming. This benchmark was created by Bixby, he describes it as a set of "real-world problems that had been collected from academic and industry sources over 21 years." Progress is based on the total time spent searching for the optimal solution for all problems in the benchmark. Progress is easy to track in this domain over this 21 year period because there were distinct releases of commercial software (CPLEX and Gurobi) that can be compared with hardware held fixed. A maximum search time of 30,000 seconds (approximately 8 hours) per problem was used, so that's what was recorded for instances where the optimum wasn't found. We clarified the trend by graphing the trend by release date rather than by version number [Bixby, 2012].

### 2.3 184x reduction in training cost (in dollars) to get to ResNet-50 performance since 2017

The eventual unit institutions generally care about for training cost is dollars. Earlier we observed a 10x efficiency improvement in terms of training FLOPs required to get ResNet-50 level accuracy (92.9% top-5 accuracy target on ImageNet). On the same target, DawnBench submissions have surpassed the contest's original benchmark cost, $2323, by a factor of 184x [Coleman et al., 2017]. This brought the cost of such a training down to $12.60 in September 2017, less than a year after the competition was announced. Training cost in dollars is a useful overall measure, that aggregates:

1. The efficiency gains from algorithmic progress we are most interested in within this paper.
2. Moore's Law's effect on GPUs, TPUs, etc.
3. Reduced cloud computing costs driven by modernization and increased competition.
4. Hardware utilization. It's not trivial to efficiently use the FLOPS capacity of GPUs, TPUs, etc.

The DawnBench results make it clear that 3. and 4. can also be notable contributions to training efficiency that are worth measuring. More targeted measurements, like training efficiency in terms of FLOPs, help clarify the takeaway from measures like DawnBench that aggregate multiple effects.

### 2.4 We can estimate costly-to-observe algorithmic efficiency improvements through scaling laws

We've focused on algorithmic efficiency improvements that are observable empirically. [Kaplan McCandlish 2020] showed that language model performance on cross-entropy had power-law scaling with the amount of compute over several orders of magnitude. Empirical scaling laws can be extrapolated to provide an estimate of how much we would have needed to scale up older models to reach current levels of performance. Through this mechanism scaling laws provide insight on efficiency gains that may require prohibitively expensive amounts of compute to observe directly.

### 2.5 Total investment in AI through private startups, public offerings, and mergers/acquisitions went up 5x between 2012 and 2018

We've primarily considered algorithmic, hardware, and data as the inputs in progress in machine learning. Money spent would be another reasonable lens since that's the lever available to decision-makers at the highest level. [Bloom et al., 2017] looks into the relationship between scientific progress and spending:

> *In many models, economic growth arises from people creating ideas, and the long-run growth rate is the product of two terms: the effective number of researchers and their research productivity... A good example is Moore's Law. The number of researchers required today to achieve the famous doubling every two years of the density of computer chips is more than 18 times larger than the number required in the early 1970s. Across a broad range of case studies at various levels of (dis)aggregation, we find that ideas – and the exponential growth they imply – are getting harder to find. Exponential growth results from large increases in research effort that offset its declining productivity.*

AI investment is also up substantially since 2012, and it seems likely this was important to maintaining algorithmic progress at the observed level. [Raymond Perrault & Niebles, 2019] notes that:

1. Private investment in AI startups rose from $7B in 2012 to $40B in 2018.
2. Investment through public offerings and mergers/acquisitions grew from $5B in 2012 to $23B in 2018.
3. The DOD is projected to invest $4.0B on AI R&D in fiscal year 2020.
4. Contract spending on AI by the US government has grown from about $150M to $728M between 2012 and 2018.

## 3 Methods

### 3.1 Main result primarily based on existing open source re-implementations of popular models

For the majority of the architectures shown in Figure 3 [Szegedy et al., 2014, Simonyan & Zisserman, 2014, He et al., 2015, Xie et al., 2016, Huang et al., 2016, Iandola et al., 2016, Zagoruyko & Komodakis, 2016, Zhang et al., 2017, Howard et al., 2017, Sandler et al., 2018, Ma et al., 2018, Tan & Le, 2019] we used PyTorch's example models [Paszke et al., 2017] with Pytorch's suggested hyperparameters. We mark our deviation from their hyperparameters in the next section. We supplemented PyTorch's example models with existing implementations of MobileNet, ShuffleNet [Xiao, 2017, Huang, 2017].

Compute used is based on the product of the following:

1. FLOPs per training image, which was counted by a PyTorch library [Zhu, 2019] that we checked against other methods for several models
2. The number of images per epoch
3. The number of epochs it took an architecture to perform better than or equal to the AlexNet model we trained

### 3.2 We made few hyperparameter adjustments between architectures and did minimal tuning

We largely followed the suggested hyperparameters from the PyTorch example models. For all points shown in figure 3 we trained using SGD with a batch size of 256, momentum of 0.9, and weight decay of 1e-4, for 90 epochs. For pre-batch norm architectures, we began with the suggested learning rate of 0.01 (GoogleNet and VGG), for all other architectures we began with the suggested learning rate of 0.1.

For AlexNet we followed the original paper's learning rate schedule of decaying by a factor of 10 every 30 epochs. For all other models, we followed the suggested 1000x total learning rate reduction. To sanity check that these were reasonable hyperparameters, we performed a scan on ResNet18 where we set the initial learning rate to 0.0316, 0.1, and 0.316 and total decay to 250x, 1000x, and 2500x. The suggested hyperparameters performed the best. For all models other than AlexNet we smoothed out the learning rate schedule, which was important for early learning as shown in Figure 2.

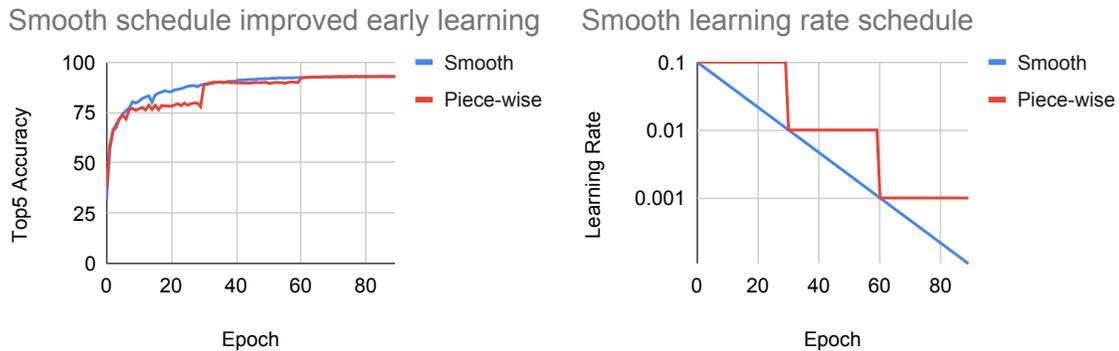

**Figure 2**   Smoothing out the learning rate improved early learning, which is the regime we were interested in. ResNet-50 learning curves pictured.

A natural concern would be that new models aren't optimized well for compute in reaching AlexNet-level performance. Before smoothing the learning rate schedule, many models hit AlexNet performance at exactly 31 epochs, when the learning rate was reduced by a factor of 10x. This adjustment often increased our measured efficiency by 2-4x, but we didn't observe meaningful differences in final performance from the change in learning rate schedule. So even though the change to the learning rate schedule could be considered minimal, it has a large effect on our measurements. The more simple shape of the updated learning curve, suggests that optimizing for convergence might be relatively compatible with optimizing for lower levels of performance, like AlexNet-level accuracy.

As context for the quality of these re-implementations we provide tables in Appendix C that compare the final accuracy we reached to the original paper results.

## 4   Results

### 4.1   Key Result: 44x less compute needed to get to AlexNet-level performance

In figure 3 we show that between 2012 and 2019 the amount of compute that neural net architectures require to be trained from scratch to AlexNet level performance has gone down by a factor of 44x (16-month doubling time)

Most researchers found the algorithmic efficiency gains to surprisingly high and regular. The progress is faster than the original Moore's Law rate (11x) over this period, where both trends made training models of AlexNet-level performance cheaper. Moore's Law is obviously a more general trend than what we observe in Figure 3. We believe it's quite interesting to see what we can say about algorithmic efficiency progress in general given these types of measurement, and we explore this question in sections 4.2 and 5.4.

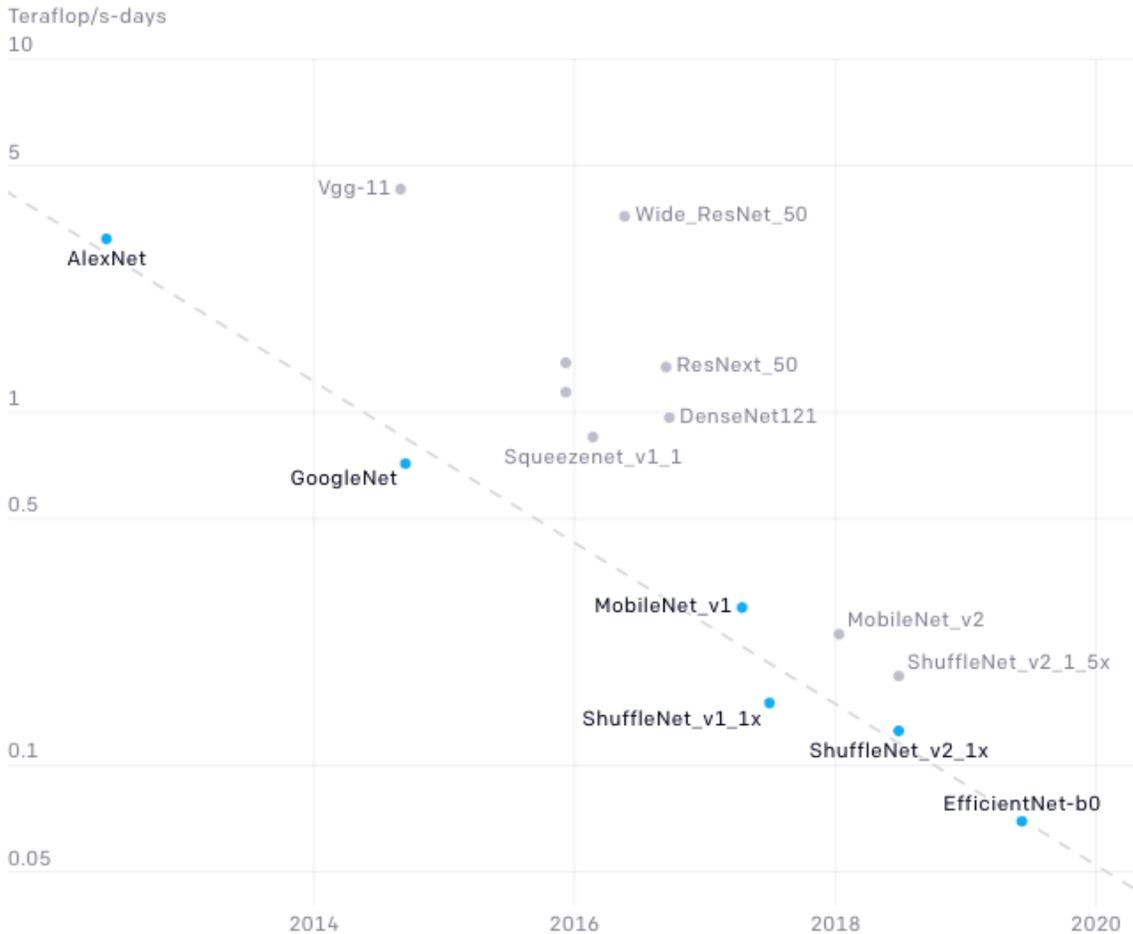

**Figure 3** Lowest compute points at any given time shown in blue, all points measured shown in gray. We observed an efficiency doubling time of 16 months.

We can split the progress in training efficiency into data efficiency (needing fewer epochs) and reductions in the number of FLOPs required per epoch. Table 1 below shows this split for the models that were the efficiency state of the art for a time.

We can see that both reductions in training epochs and FLOPs per training image play an important and varying factor in the overall algorithmic efficiency gains. This type of analysis is somewhat sensitive to how far the original work pushed towards convergence.[2] Other limitations are discussed in sections 5.4 and 5.7. Calculations for the figure 3 are provided in Appendix B. Relevant information for EfficientNet training cost was provided through correspondence with authors.

---

[2] It only took 62 of the 90 epochs for AlexNet to train to 78.8% top 5 accuracy on ImageNet (99.6% of the 79.1% final accuracy). So if the original AlexNet had only been trained for 62 epochs, we would have calculated the overall algorithmic efficiency gain as 30x rather than 44x. We don't think it's tractable to mitigate this confounder without adding a lot of complexity to explaining the measurement, but it seemed important to flag as a limitation of our approach.

**Table 1** Breakdown of total training efficiency gains in reaching AlexNet-level accuracy into reduction of training epochs and flops per epoch

| Experiment | Training epochs factor | FLOPs per epoch factor | Training efficiency factor |
|---|---|---|---|
| AlexNet | 1.0 | 1.0 | 1.0 |
| GoogleNet | 11 | 0.38 | 4.3 |
| MobileNet_v1 | 8.2 | 1.35 | 11 |
| ShuffleNet_v1_1x | 3.8 | 5.5 | 21 |
| ShuffleNet_v2_1x | 4.5 | 5.5 | 25 |
| EfficientNet-b0 | 22 | 2.0 | 44 |

### 4.2 FLOPs based learning curves can help clarify comparisons between models

We find it noteworthy that in when we plot FLOPs based learning curves in figure 4 some architectures dominate others.

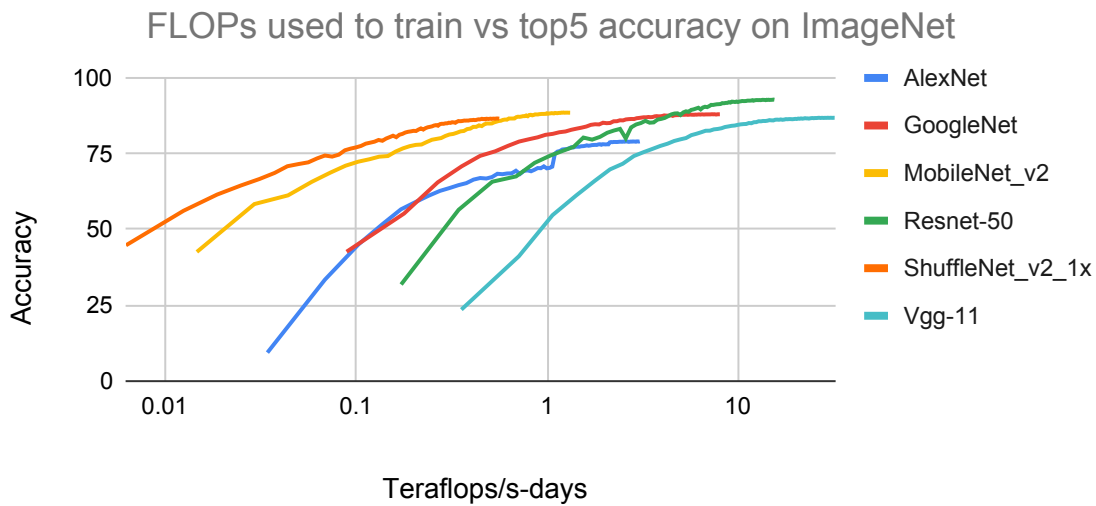

**Figure 4** Some models reach all levels of of accuracy using less compute than other models

FLOPs based learning curves can help clarify what type of advances a new architecture consists of. ResNet-50 dominates VGG-11 and GoogLeNet dominates AlexNet on this plot. That is for all amounts of training compute they get better accuracy. VGG-11 reached higher final accuracy than AlexNet, but it took more compute to get to all levels of performance than AlexNet.

### 4.3 We observed a similar rate of progress for ResNet-50 level classification performance and faster rates of efficiency improvement in Go, Dota, and Machine Translation

We're also interested in measuring progress on frontier AI capabilities, the capabilities that are currently attracting the most attention and investment. It seems to us as if language modeling [Devlin et al., 2018, Radford et al., , Raffel et al., 2019] and playing games [Silver et al., 2016, Silver et al., 2017, Silver et al., 2018, OpenAI et al., 2019] are the domains of interest given our criteria.

Within those domains, our desiderata were:

1. task of sufficient difficulty to demonstrate that improvements work at scale [Sutton, 2019]

2. benchmark of high interest over long horizon in which there's general agreement we've observed large progress in capabilities.
3. sufficiently good publicly available information/re-implementations to easily make an estimate

It's hard to get all these desiderata, but Table 2 below summarizes all the data we have observed.

Table 2 Increased efficiency (in terms of FLOPs) in reaching the same performance on select tasks.

| Original | Improved | Task | Efficiency Factor | Period | Doubling Time |
|---|---|---|---|---|---|
| AlexNet | EfficientNet | ImageNet | 44x | 6 years | 16 months |
| ResNet | EfficientNet | ImageNet | 10x | 4 years | 17 months |
| Seq2Seq | Transformer | WMT-14 | 61x | 3 years | 6 months |
| GNMT | Transformer | WMT-14 | 9x | 1 year | 4 months |
| AlphaGo Zero | AlphaZero | GO | 8x* | 1 year* | 4 months* |
| OpenAI Five | OpenAI Rerun | Dota | 5x* | 2 months* | 25 days* |

*The work on Go and Dota are over shorter time scales and more the result of one research group rather than a large scientific community, so those rates of improvement should be considered to apply to a different regime than the rates in image recognition and translation.

When we apply this lens to translation [Sutskever et al., 2014, Vaswani et al., 2017] it shows more progress than vision over a shorter time horizon. Though we only have short horizon progress for Go and Dota, we'd only need to see a modest 3x and 5x efficiency gain over 5 years for their rates to surpass the rate of progress on the vision task. The underlying calculations are provided in appendix A.

One might worry that the rate of progress in image recognition is very sensitive to performance level chosen, so we also did a shallow investigation of efficiency gains at ResNet-50 level of performance. The relevant information, that EfficientNet-b0 took 4 epochs to get to AlexNet level accuracy, and EfficientNet-b1 [Tan & Le, 2019] took 71 epochs to get to ResNet-50 level accuracy was provided through correspondence with authors (where each was trained with 1 epoch of warmup rather than 5).

**We observed a similar rate of progress for efficiency gains in inference on ImageNet.** We also did a shallow investigation into how the rate of progress on inference efficiency has compared to training efficiency. We observed that:

1. Shufflenet [Zhang et al., 2017] achieved AlexNet-level performance with an 18x inference efficiency increase in 5 years (15-month doubling time).
2. EfficientNet-b0 [Tan & Le, 2019] achieved ResNet-50-level performance with a 10x inference efficiency increase in 3 and a half years (13-month doubling time).

These results suggest that training efficiency and inference efficiency might improve at somewhat similar rates. Though it's important to note we have many fewer points across time and domains for inference.

## 5 Discussion

### 5.1 We attribute the 44x efficiency gains to sparsity, batch normalization, residual connections, architecture search, and appropriate scaling

A more thorough study would have carefully ablated all the features of interest from successful models while controlling for model size to be able to attribute the efficiency gains to specific improvements in a quantitative manner [Lipton & Steinhardt, 2018]. We performed some ablations, but primarily rely on less direct evidence when forming opinions about which improvements we suspect were most important to the 44x increase in efficiency. For instance we discuss what the original authors credit, though it's important to recognize authors are incentivized to emphasize novelty. We think it's important to note that efficiency gains may compose in a hard to predict, non-linear manner.

**Batch Normalization:** Batch normalization enabled a 14x reduction in the number of floating-point operations needed to train to Inception level accuracy [Ioffe & Szegedy, 2015]. It's unclear how such algorithmic efficiency gains like batch normalization compose, but it seems reasonable to attribute some meaningful portion of the gains to normalization. We made a few attempts to try and train a ShuffleNet without batch normalization, but we were unable to get a model to learn. We suspect we would have needed to carefully initialize the network to do so [Zhang et al., 2019].

**Residual Connections:** ShuffleNet units, the building blocks of ShuffleNet, are residual blocks. EfficientNet also has residual connections.

**Sparsity:** GoogLeNet was explicit in describing sparsity as the primary inspiration for its architecture, and GoogLeNet alone was a 4.3x efficiency improvement over AlexNet. [Szegedy et al., 2014].

> *This raises the question of whether there is any hope for a next, intermediate step: an architecture that makes use of the extra sparsity, even at filter level, as suggested by the theory, but exploits our current hardware by utilizing computations on dense matrices.*

ShuffleNet largely credits replacing dense 1 x 1 convolutions with a sparser structure. If we assume all the ShuffleNet gains came from sparsity, batch normalization, and residual connections, it seems reasonable to credit sparsity with being able to produce at least the 4.3x that came with GoogLeNet (leaving 5.8x of the 25x gain shown in Table 1 for the other two conceptual improvements).

**Appropriate Scaling:** Given it's architecture AlexNet was optimally sized for AlexNet-level performance. Given our tests of scaled up and scaled down models ShuffleNet_v2_1x, and EfficientNet-b0 seem to be close to appropriately sized for AlexNet-level performance. We tested the effect of scaling by scaling down a ResNet-50 by EfficientNet's compound scaling factor twice (1.4x less depth, 1.2 less width, 1.3 lower resolution) [Tan & Le, 2019]. Scaling the ResNet architecture to a more appropriate size for AlexNet-level performance yielded a 2.1x improvement in algorithmic efficiency for AlexNet-level performance. Figure 8 in the EfficientNet paper shows that their compound scaling techniques (systematically scaling width, depth, and resolution) can result in 5x or more gains in algorithmic efficiency over more naive scaling approaches.

**Architecture Search:** EfficientNet seems to attribute much of its improved performance to leveraging architecture search rather than iterating on hand designed architectures. EfficientNet was a 1.8x increase in algorithmic efficiency over ShuffleNet at AlexNet-level performance.

### 5.2 It's unclear the degree to which the observed efficiency trends generalize to other AI tasks

We're most interested in what our small number of data points suggest about algorithmic progress overall during this period. We recognize it's difficult to go from one or more specific measures to stating anything about overall progress. In this section we share our current impressions and suggest measures that could clarify the degree to which the trends we've observed generalize.

All our measures were for tasks that have:

1. received large amounts of investment (researchefr time and/or compute)
2. in which there's general agreement we've observed large progress in capabilities.

We suspect that this style of measurement on tasks that meet these criteria is likely to show similar rates of improvement in algorithmic efficiency as we've observed here. One concern we had, was that the rates of improvement would be very dependent on the level of performance. That may still be the case, but we were surprised how close the efficiency doubling time was for AlexNet-level performance (16 months) and ResNet50-level performance (17 months). We also suspect, but are less confident, that such measurements would should similar progress in these domains (image recognition, natural language processing, and games). We'd be very interested in such measurements.

However, we're also interested in progress in high potential tasks that don't fit these criteria, like certain reasoning tasks. In the previous section, we attributed the efficiency gains over AlexNet primarily to sparsity, residual connections, normalization, principled scaling, and architecture search all of which are relatively task-agnostic. But, it's possible that we'd observe only small efficiency gains from these techniques on such tasks. We consider the degree to which the observed efficiency trends generalize to other AI tasks a highly interesting open question.

## 5.3 Why new capabilities are probably a larger portion of progress than observed efficiency gains

AlexNet achieved performance that no system had previously achieved. We can try to reason about how much compute would have been required in scaling up previous systems to match AlexNet's performance. From this point of view, we believe AlexNet represented significant progress in how much compute was required to achieve AlexNet-level performance. This analysis doesn't attempt to quantify that progress because it's less tractable. More generally, the first time a capability is created, algorithmic breakthroughs may have been leveraged to dramatically reduce the resources that would have otherwise been needed. For instance, if we imagine simply scaling up a DQN [Mnih et al., 2013] model to play Go it could easily have needed 1000x or more times as much compute to reach AlphaGo level. Such efficiency gains are not generally observed empirically, though they can be calculated with asymptotic analysis in some cases and estimated with empirical scaling laws in others [McCandlish et al., 2018].

More formally, if we go far enough back in time, algorithmic progress takes us from brute force search to lower complexity classes, which is what enables capabilities of interest to be built at all. Within this zoomed-out view, the progress that went into making a capability possible at all, in total, yields an astronomically larger algorithmic efficiency improvement factor than directly observed efficiency improvements for capabilities that have recently been observed for the first time. This limit analysis lends some support to the claim that the rate of gain in algorithmic efficiency on a capability of interest might often be faster before a capability is observed.

In the DQN and brute force examples described above, we find it most helpful to start by thinking of a scaling law, a plot of performance vs training compute used. Our algorithmic efficiency data results are points we find meaningful from those graphs, but sometimes similar comparisons would just yield an astronomical number that might not have much meaning. In such cases, we'd recommend analyzing a graph of the scaling law, since it contains the entire picture.

While most researchers we've discussed the result with found the 44x number surprisingly high, because of this effect 44x may strongly underestimate algorithmic progress on image classification during this period. When this analysis is discussed in the context of the relative importance of advancements in hardware and software in AI progress, we think it's critical to remember this limitation [Sutton, 2019].

## 5.4 We estimate a 7.5 million times increase in the effective training compute available to the largest AI experiments between 2012 and 2018

This section explains why we estimate there was a 7.5 million times increase in the effective training compute (in FLOPs) available to the largest AI experiments during this period. The reasoning behind our estimate is that's what we get when we take the product of the AI and Compute trend [Amodei & Hernandez, 2018] (300,000x) and AlexNet efficiency trend found in this work (25x over this period[3]), and carefully consider what this product means. When we consider that we have more compute and that each unit of compute can do more, it becomes clear that these two trends are somehow multiplicative.

This section is more speculative than the rest of the paper, but we think it's important to explore the potential implications of our efficiency measurements. We believe a 7.5 million times estimate is somewhat defensible when we:

1. Narrowly define capabilities of interest so that 300,000x can be applied by definition.
2. Define what we mean by effective compute.
3. Discuss major considerations for why 25x could be an underestimate/overestimate for algorithmic progress on capabilities of interest.

**Capabilities of interest:** We define capabilities of interest as the training runs at close to the peak of size that was observed in 2018. Therefore it's appropriate to apply the 300,000x from AI and Compute trend by definition. By 2020 such systems include AlphaZero, OpenAI Five, and NLP systems. This definition helps us avoid having to reason about what our measurements imply for distant domains. We have some measurements of progress for many of the capabilities of interest by the above definition. Though it's possible there are unpublished results that fit the capability of interest definition in relatively distant domains.

---

[3]Through 2018 we use the 25x efficiency gains ShuffleNet represented rather than the 44x gains that EfficientNet represented in 2019

**Effective compute:** The conception we find most useful is if we imagine how much more efficient it is to train models of interest in 2018 in terms of floating-point operations than it would have been to "scale up" training of 2012 models until they got to current capability levels. By "scale up," we mean more compute, the additional parameters that come with that increased compute, the additional data required to avoid overfitting, and some tuning, but nothing more clever than that. We considered many other conceptions we found less helpful [4].

**Why our overall take is that 25x is likely an underestimate for algorithmic progress on capabilities of interest**  Our overall take relies heavily on our observations in the domain of interest. We saw larger overall progress in NLP and faster rates of short horizon progress for Go and Dota. In NLP we observed a 60x efficiency factor over 3 years for machine translation. Though we only have short-horizon progress for Go and Dota, we'd only need to see a modest 3x and 5x efficiency gains respectively over 5 years for their rates to surpass the rate of progress on the vision task.

On the other hand, algorithmic progress has a domain specific component, and it's unclear how representative the 25x is of the average efficiency progress in the broader domain of AI during this period. However, we believe this effect is smaller than the effect in the opposite direction of not measuring the contribution of new capabilities like AlexNet, Seq2Seq, or original AlphaGo systems during this period. In section 5.3 we provided arguments for why new capabilities might represent 100x or more algorithmic efficiency progress.

To further clarify what drove changes in effective compute over this period, we split the AI and Compute trend into Moore's Law and increased spending/parallelization[5]. We graph an estimate for the effective compute trends in terms of these two components as well as progress in algorithmic efficiency in figure 5 below.

**We're uncertain whether hardware or algorithmic progress actually had a bigger impact on effective compute available to large experiments over this period**, because of the ways we've discussed in which the algorithmic estimate is conservative. Most researchers found the algorithmic efficiency progress to be surprisingly fast. So, regardless of one's interpretation of what the AI and Compute trend implies about future AI progress, we believe our algorithmic efficiency estimates suggests:

1. a modest update towards expecting faster progress along the edge of what's possible for AI to do in the short term.

2. potentially large update on long term expectations about AI if the algorithmic efficiency on capabilities of interest continues to improve at a similar rate.

**Directly commenting on the likelihood of any of the 3 macro trends in figure 5 continuing in the future is out of scope for this work.** Making credible forecasts on such topics is a substantial enterprise, we'd rather avoid here than give insufficient treatment. Rather we present the evidence we see as relevant for a reader who'd like to form their own expectations about extrapolating trends in algorithmic efficiency.

**Additional reasons why 44x over 7 years could be an underestimate for progress on AlexNet-level algorithmic efficiency:**

1. Only AlexNet was heavily optimized for AlexNet level performance. Models are generally tuned for performance at convergence, not early learning. Our results were produced with minimum tuning for early learning and AlexNet-level performance, and tuning them could only increase their efficiency gains.

2. It's our understanding that the re-implementation of AlexNet we used had a better initialization scheme than the original work. This effect adds another factor of conservativeness to our analysis. We expect future analysis to also be limited by this effect. This concern could be mitigated by researchers publishing their learning curves in addition to training compute used to train.

3. We don't account for gains from being able to use lower precision computation [Gupta et al., 2015].

4. We don't account for gains from increased GPU utilization or improved GPU kernels.

---

[4] Our initial thinking was in terms of what an elite team in 2012 could have done if given a large amount compute, but this was unobservable. We could make something similar observable by having a group of smart physicists/mathematicians that were unfamiliar with modern ML methods work on problems without access to modern results, but that would be very expensive to observe.

[5] Increased spending and parallelization are coupled in that given fixed time a researcher is limited by both (i) how many concurrent GPU's are available to them which is primarily a financial question, and (ii) how many GPU's can productively be applied to the problem, which is a scientific question [McCandlish et al., 2018, Jia et al., 2018]

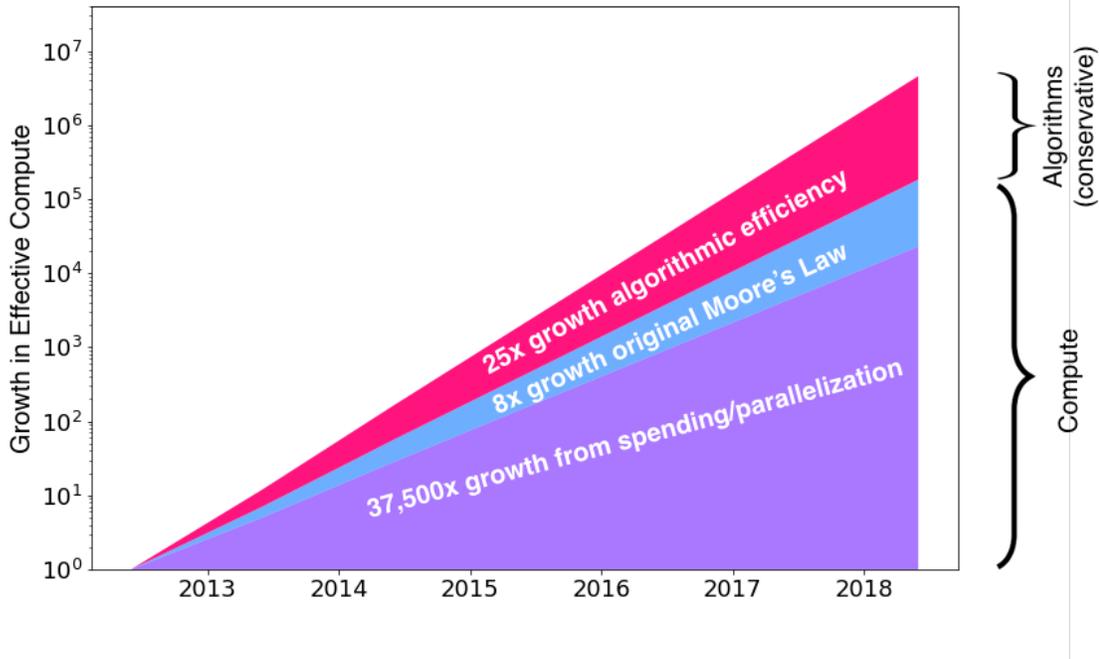

**Figure 5** The notion of effective compute allows us to combine AI and Compute trend and this result in a single graph. These trends multiply as in addition to being able to do more with a fixed amount of compute now, researchers have more of it. The AI and Compute trend is decomposed into a hardware efficiency gain estimate (original Moore's Law) and money/parallelization [Moore, 1965, Amodei & Hernandez, 2018]. This estimate, as discussed in the body of this section, is more speculative than the rest of the paper, but we think it's important to explore the potential implications of our efficiency measurements.

### 5.5 It's possible there's an algorithmic Moore's Law for optimization problems of interest

This work suggests that in high investment areas of AI algorithmic efficiency improvement is currently having a similar-sized effect as Moore's Law has had on hardware efficiency in the past. Others have noticed comparable algorithmic progress over decades in related domains like Chess, Go, SAT solving, and operations research. In light of that past analysis, it's less surprising that we've observed algorithmic efficiency gains this large on training to an AlexNet level of performance. The common thread here seems to be that these along with AI systems are all optimization problems of interest.

Systematic measurement could make it clear whether an algorithmic equivalent to Moore's Law in the domain of AI exists, and if it exists, clarify its nature. We consider this a highly interesting open question. We suspect we're more likely to observe similar rates of efficiency progress on similar tasks. By similar tasks we mean within these sub-domains of AI, wide agreement of substantial progress, and comparable levels of investment (compute and/or researcher time). It's also unclear the degree to which general vs domain specific gains would be the drivers of such progress, and how gains compound over long periods as the field progresses through several benchmarks. Problems of high investment might be be quite biased towards ones we're making progress on rather, where an ideal measure might focus on the questions that are seen as most important.

An AI equivalent to Moore's Law would be harder to measure, because it's not about progress on a single problem, it's about progress on the frontier of optimization problems. Through that lens, it seems more plausible we'll see long term exponential progress on algorithmic efficiency for AI capabilities of interest if our primary finding is an extension of an existing, long-running trend in progress on optimization problems of interest.

## 5.6 Research provides leading indicators of the future economic impact of AI

The eventual overall measure of AI research's impact on the world will likely be economic. However, it took past general-purpose technologies like electrification and information technology a surprisingly long time to become widespread. From the start of information technology era it was about 30 years before personal computers were in more than half of US homes [Jovanovic & Rousseau, 2005] (similar timeline for personal computers). Analysis of past investments in basic research along 20-30 year timescales in domains like computers indicates that there's at least some tractability in foreseeing long term downstream impacts of technology like machine learning. Economic trends of AI are very informative, but measures of research progress are of particular interest to us as leading indicators of the eventual downstream economic and societal impact.

## 5.7 Major limitations

The limitations of this work are discussed throughout, but the major ones are reiterated here:

1. We only have a small number of algorithmic efficiency data points on a few tasks (Section 4). It's unclear the degree to which we'd expect the rates of progress we've observed to generalize to algorithmic efficiency progress on other AI task and domains. We consider this a highly interesting open question that we discuss in Section 5.2.

2. We believe our approach underestimates algorithmic progress, primarily because new capabilities are likely a larger portion of algorithmic progress than observed efficiency gains (Section 5.3). This weakness could be addressed by fitting scaling laws to estimate the cost of prohibitively expensive training runs (Section 2.4).

3. This analysis focuses on the final training run cost for an optimized model rather than total development costs. Some algorithmic improvements make it easier to train a model by making the space of hyper-parameters that will train stably and get good final performance much larger. On the other hand, architecture searches increase the gap between the final training run cost and total training costs. We believe a quantitative analysis of these effects would be very informative, but it's beyond the scope of this paper.

4. We don't comment on the degree to which we believe efficiency trends will extrapolate, we merely present our results (Section 4) and the related work (Section 2) we think is relevant for someone attempting to make such a prediction. Though we do comment on the implications if the trends persist.

# 6 Conclusion

We observe that hardware and algorithmic efficiency gains multiply and that neither factor is negligible over meaningful horizons, which suggests that a good model of AI progress should integrate measures from both.

We hope this work is helpful to those trying to understand, measure, and forecast AI progress in a variety of settings. We've observed that AI models for high interest tasks are getting cheaper to train at an exponential rate faster than Moore's Law. Even though we're early on in applying this trend to AI, we were surprised and inspired to learn that the original Moore's Law was coined when integrated circuits had a mere 64 transistors (6 doublings) [Moore, 1965] and naively extrapolating it out predicted personal computers and smartphones (an iPhone 11 has 8.5 billion transistors). If we observe decades of exponential improvement in the algorithmic efficiency of AI, what might it lead to? We're not sure. That these results make us ask this question is a modest update for us towards a future with powerful AI services and technology. Conversely, if we were to start only observing incremental gains (say 2x improvements every 5 years), we think that'd be a meaningful and widely understandable indicator that algorithmic progress had slowed down.

More ambitiously, we hope that reporting on algorithmic efficiency improvements will become a strong and useful norm in the AI community. Improved performance is what AI algorithms are ultimately judged by. Algorithmically efficient models on benchmarks of interest are promising candidates for scaling up and potentially achieving overall top performance. Efficiency is straightforward to measure, as it's just a meaningful slice of the learning curves that all experiments generate. Given these considerations and the primacy of efficiency in measuring progress in computer science, we believe there's a strong case for reporting on and tracking training efficiency states of the art over time.

# 7 Acknowledgements

We'd like to thank the following people helpful conversations and/or feedback on this paper: Dario Amodei, Jack Clark, Alec Radford, Paul Christiano, Sam McCandlish, Ilya Sutskever, Jacob Steinhardt, Jared Kaplan, Amanda Askell, John Schulman, Ryan Lowe, Tom Henighan, Jacob Hilton, Asya Bergal, Katja Grace, Ryan Carey, Nicholas Joseph, and Geoffrey Irving.

Thanks to Niki Parmar for providing the relevant points from the transformer learning curves [Vaswani et al., 2017].

Also thanks to Mingxing Tan for providing the relevant points from EfficientNet learning curves and running an experiment with reduced warmup [Tan & Le, 2019].

## A  Calculations for efficiency improvements in Go, Dota, and Machine Translation

**Machine Translation:**  We estimate that the Transformer [Vaswani et al., 2017] required 61x less compute to get to Seq2Seq-level of performance [Sutskever et al., 2014] on English to French translation on WMT'14 3 years later. This estimate is based on:

1. total training compute used by the transformer base model in original paper (3.3e18 FLOPs)
2. compute estimate for Seq2Seq in AI and Compute (4.0e19 FLOPs)
3. the base transformer got to Seq2Seq level around 20% of the way through it's run. (provided by authors of transformer paper).

$$4.0e19/(0.20 * 3.3e18) = 61$$

We estimate the the Transformer [Vaswani et al., 2017] required 9x less compute to get to GMNT-level of performance on English to French translation on WMT-14 1 year later. This estimate is based on:

1. total training compute used by the transformer big model in original paper (2.3e19 FLOPs)
2. compute estimate for GMNT transformer paper (1.4e20 FLOPs)
3. the base transformer got to Seq2Seq level around 68% of the way through it's run. (provided by authors of transformer paper).

$$1.4e20/(0.68 * 2.3e19) = 9$$

**AlphaGo Zero to AlphaZero:**  We estimate that AlphaZero [Silver et al., 2018] required 8x less compute to get to AlphaGo Zero [Silver et al., 2017] level approximately one year later. We don't currently have enough information to compare to AlphaGo Lee [Silver et al., 2016].This is based on:

1. an estimated 4.4x decrease in total FLOPs used to train AlphaZero in AI and Compute
2. it took AlphaZero 390,000 of the 700,000 steps it was trained for to match AlphaGo Zero performance.

$$4.4 * (700,000/390,000) = 8$$

**OpenAI Five Rerun:** OpenAI Five "Rerun" got to the same skill level from scratch on the final environment without surgery using 5x less compute 2 months after the OG match [OpenAI et al., 2019]. However, some hard to pin portion of the additional cost came from a changing environment, as there were balance change patches approximately every 2 weeks during the original 10 month training period.

## B  Calculations for efficiency improvements in image classification

Table 3  FLOPs required to reach same AlexNet level accuracy

| teraflop/s-days | Experiment | Epochs | gigaflops/img (used) | gigaflops/img (THOP) | gigaflops/img (paper) |
|---|---|---|---|---|---|
| 367.7 | Vgg-11 | 12 | 7.98 | 7.98 | - |
| 308.0 | Wide_ResNet_50 | 7 | 11.46 | 11.46 | - |
| 266.1 | AlexNet | 90 | 0.77 | 0.77 | - |
| 118.6 | Resnet-50 | 8 | 3.86 | 3.86 | - |
| 118.5 | Resnet-34 | 9 | 3.43 | 3.43 | - |
| 115.3 | ResNext_50 | 7 | 4.29 | 4.29 | - |
| 97.9 | Resnet-18 | 15 | 1.70 | 1.70 | - |
| 82.9 | DenseNet121 | 8 | 2.70 | 2.70 | - |
| 73.1 | Squeezenet_v1_1 | 53 | 0.36 | 0.36 | - |
| 61.4 | GoogLeNet | 8 | 2.00 | 2.00 | - |
| 24.0 | MobileNet_v1 | 11 | 0.57 | 0.58 | 0.57 |
| 20.2 | MobileNet_v2 | 16 | 0.33 | 0.33 | - |
| 15.4 | ShuffleNet_v2_1_5x | 13 | 0.31 | 0.31 | - |
| 12.9 | ShuffleNet_v1_1x | 24 | 0.14 | 0.15 | 0.14 |
| 10.8 | ShuffleNet_v2_1x | 20 | 0.14 | 0.15 | 0.14 |
| 6.0 | EfficientNet-b0 | 4 | 0.39 | - | 0.39 |

Where $training\_flops = epochs * flops\_per\_image * images\_per\_epoch$
With $images\_per\_epoch = 1.28 * 10^6$ and a $teraflop/s - day = 1e12 * (24 * 60 * 60 s/day)$

## C  Accuracy achieved in relevant models

Table 4  Top-5 final training accuracy comparisons for relevant models

| Experiment | My Top-5 | Pytorch/Examples Top-5 | Paper Top-5 | Single Crop Validation* |
|---|---|---|---|---|
| AlexNet | 79.0% | 79.1% | 83.0% | ? |
| Vgg-11 | 86.8% | 88.6% | 93.0% | no |
| GoogLeNet | 88.0% | 89.5% | 89.9% | yes |
| Resnet-50 | 92.8% | 92.9% | 93.3% | yes |
| Squeezenet_v1_1 | 80.6% | 80.6% | 80.3% | ? |

Table 5  Top-1 final training accuracy comparisons for relevant models

| Experiment | My Top-1 | Pytorch/Examples Top-1 | Paper Top-1 | Single Crop Validation* |
|---|---|---|---|---|
| MobileNet_v1 | 71.0% | - | 70.6% | yes |
| MobileNet_v2 | 68.5% | 71.9% | 72.0% | yes |
| ShuffleNet_v1_1x | 64.6% | - | 67.6% | yes |
| ShuffleNet_v2_1_5x | 69.3% | 69.4% | 71.6% | yes |

*We use a single center 224x224 crop for evaluating performance on the validation data points for all of our models, but not all of original papers evaluate performance in this manner.